# A character representation enhanced on-device Intent Classification

**Sudeep Deepak Shivnikar, Himanshu Arora, Harichandana B S S**
Samsung R & D Institute, Bangalore
{s.shivnikar, him.arora, hari.ss}@samsung.com

## Abstract

Intent classification is an important task in natural language understanding systems. Existing approaches have achieved perfect scores on the benchmark datasets. However they are not suitable for deployment on low-resource devices like mobiles, tablets, etc. due to their massive model size. Therefore, in this paper, we present a novel light-weight architecture for intent classification that can run efficiently on a device. We use character features to enrich the word representation. Our experiments prove that our proposed model outperforms existing approaches and achieves state-of-the-art results on benchmark datasets. We also report that our model has tiny memory footprint of ~5 MB and low inference time of ~2 milliseconds, which proves its efficiency in a resource-constrained environment.

## 1 Introduction

In a time where consumers and businesses alike are constantly adopting new technologies in hope of increasing efficiency and convenience, the intelligent virtual assistant (IVA) has been an immediate success[1]. For a high-quality IVA, it is very crucial to understand the intentions behind customer queries, emails, chat conversations, and more in order to automate processes and get insights from customer interactions. Thus research interest in Intent detection is on the rise.

Intent classification is an important task in Natural Language Understanding (NLU) systems which is the task of assigning a categorical intent label to an input utterance. Most IVAs use cloud-based solutions and mainly focus on accuracy rather than model size. But due to factors like privacy and personalization, there is a need for deploying models on device and thus on-device intent classification is significant. The current state-of-the-art models are highly accurate on benchmark datasets. However, most of these models have a huge number of parameters and use complex operations. Due to these reasons, they are not suitable for deployment on low-resource devices like mobiles, tablets, etc.

Gartner [2] predicts that by 2020, 80% of the smartphones shipped will have on-device AI capabilities. For this, there is a need for light-weight, fast and accurate models that can run efficiently in a resource-constrained environment. Thus, in this paper, we propose an on-device intent classification model.

In our proposed model, we use character features along with word embeddings to get enriched word representations. We use Long Short Term Memory Recurrent Neural Network (LSTM-RNN) (Hochreiter and Schmidhuber, 1997) to obtain the context vector for the input utterance. Further, we benchmark our model against publicly available ATIS and SNIPS datasets. Our experiments show that the use of character features has resulted in improved accuracy on the benchmark datasets.

The major contributions of this paper are given below.

- We propose a novel on-device architecture for Intent Classification which uses character features along with word embeddings.
- We benchmark our model against publicly available ATIS and SNIPS

---

[1] https://www.statista.com/topics/5572/virtual-assistants/

[2] https://www.gartner.com/en/newsroom/press-releases/2018-03-20-gartner-highlights-10-uses-for-ai-powered-smartphones

- datasets and achieve state-of-the-art results.
- We measure the system-specific metrics like RAM usage and inference time and show that our proposed model is efficient for low-resource devices.

The rest of the paper is organized as follows. In section 2, we give a brief overview of existing approaches for the task. We describe our approach in detail in section 3. Experimental results are presented in section 4. In section 5, we conclude and discuss future work.

## 2 Related work

Intent classification is the task of predicting an intent label for the given input text. It is a well-researched task. Early research include maximum entropy Markov models (MEMM) by Toutanova and Manning (2000). Haffner et al. (2003) and Sarikaya et al. (2011) have approached this task using Support Vector Machines (SVM).

Another popular model used were CRF based methods. Lafferty et al. (2001) first proposed Conditional (CRF) to build probabilistic models for segmentation and labelling sequence data which was proved to perform better over MEMMs. Following this, Triangular-chain conditional random fields was proposed by Jeong and Lee (2008) which is used to jointly represent the sequence and meta-sequence labels in a single graphical structure. This method outperformed the base model.

Purohit et al. (2015) have demonstrated the effectiveness of using knowledge-guided patterns in short-text intent classification. Sridhar et al (2019) have proposed the use of semantic hashing for intent classification for small datasets.

Recently joint models for intent classification and slot filling have been developed. Taking inspiration from TriCRF, Xu and Sarikaya (2013) proposed a CNN-based TriCRF for joint Intent and Slot filling. This was a neural network version of TriCRF which outperformed the base model by 1% for both intent and slot. A joint model using Gated Recurrent Unit (GRU) and max pooling for intent detection and slot filling was developed by Zhang and Wang (2016). Following this, Hakkani-Tur et al. (2016) and Liu and Lane (2016) also developed a joint model using recurrent neural networks. To model the relationship between the intent and slots, Goo et al. (2018) and Li et al. (2018) have used gate mechanism. Wang et al. (2018) have proposed Bi-model based RNN semantic frame parsing network structures by considering the cross-impact of both the tasks. Zhang et al. (2019) have used capsule networks that considers the hierarchical relationships between words, slots, and intents. E et al. (2019) have used SF-ID network to provide bidirectional interrelated mechanism for intent detection and slot filling tasks. Qin et al (2019) have used stack-propagation framework to better model the relationship between slots and intents. They further use BERT with their approach to achieve the current state-of-the-art results.

Although above mentioned joint models achieve impressive results on benchmark datasets, they are inefficient for the applications where only intent information is sufficient. Also, their heavy architecture and large model size make their on-device deployment difficult. Most of these models use multiple layers of operations that result in higher RAM usage and inference time. Our proposed model is light-weight, fast, and accurate, which makes it highly efficient for deployment on low-resource devices.

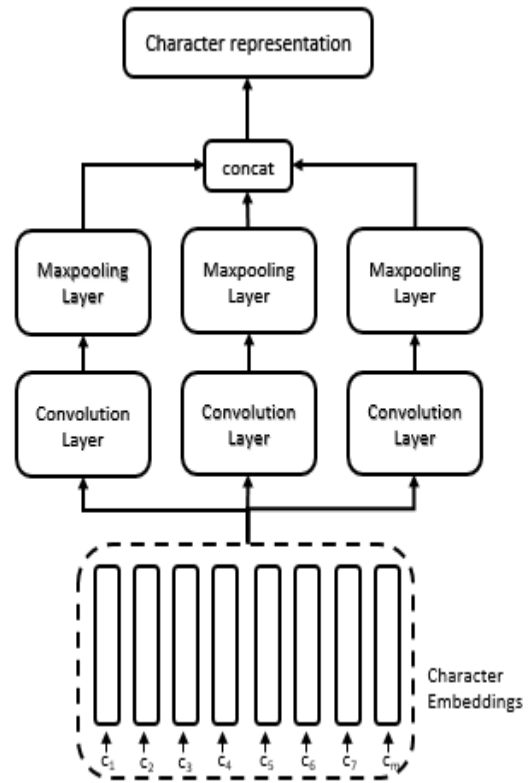

Figure 1: Architecture of character feature extractor

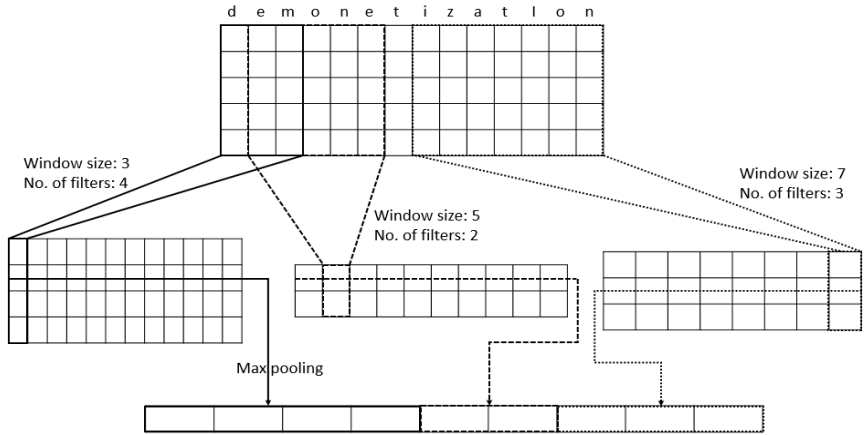

Figure 2: Illustration of character feature extraction.

## 3 Approach

In this section, we discuss our approach in detail. We use a light-weight architecture that can be efficiently deployed on device. We use character-level representation along with word embedding to enrich word-level representation.

**Character representation**: The use of character-level features to represent a word has proven useful for multiple NLP tasks. It has been used for language modeling (Kim et al., 2015), parts of speech (POS) tagging (Santos and Zadrozny, 2014), named entity recognition (NER) (Santos and Guimarães, 2015), etc. The use of character level features makes the model robust towards spelling mistakes. Since representations are formed using characters, out-of-vocabulary (OOV) words also get representation which can be further fine-tuned. It also helps to get similar representations for words with common root/prefix. For example consider the following three words: petrify, petrifies, and petrifying. These three words get similar representation using character features as they share a common prefix 'petrif'.

The architecture used to get character representation is depicted in Figure 1. Each word is a sequence of characters. Character embeddings are used to encode this information. Character embeddings are initialized randomly and learned during training. These character embeddings are fed to 3 convolution layers. Convolution layers have different convolution windows which help them to capture different character features. Max-pooling is performed on the output of convolution to select dominant features. The output of max-pooling layers is concatenated to get the character level representation.

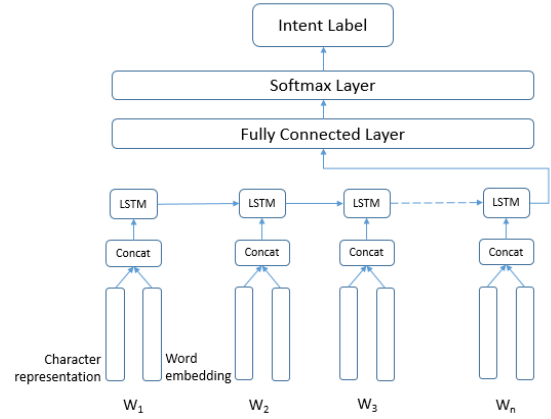

Figure 3: Architecture of proposed model

An illustration of the above-mentioned process is shown using figure 2. For illustration, we use character embedding size of 5. Convolution window sizes and filter sizes are set to (3, 5, 7) and (4, 2, 3) respectively. Max-pooled vectors are concatenated to get character features.

**Word level representation**: We use word embeddings to capture semantic information. They are initialized with pre-trained embeddings and fine-tuned during training. Pre-trained word embeddings are trained on a huge corpus and hence capture the semantic representation of the word well, which can be further fine-tuned. The use of pre-trained word embeddings helps the model to converge quickly resulting in lower training time. We use pre-trained glove embeddings (Pennington et al., 2014). Final word-level representation is obtained by concatenating word embedding with the character-level representation of the word. These concatenated embeddings are fed as input to the encoder.

| Intent | Example |
|---|---|
| wish | Wish you a very happy birthday James! |
| invitation | You are invited to our wedding. Please attend. |
| announcement | We are going to be parents! |
| love | Love you to the moon and back dear! |
| thank | Thank you for your unconditional love and useful advice! |
| miss | I hope to see you pretty soon as I miss you way too much, dear. |
| sorry | I hope you can accept my apology and get rid of my guilt. Sorry |
| job posting | We are hiring! A Registered Pharmacist, is needed at our Pune office |
| sale | Get 39% off on Super Skinny Women Blue Jeans. Hurry up Stock is limited |
| quotes | The more you fall, the more stronger you become for getting up. Never give up no matter what. |

Table 2: Details of custom datasets.

An encoder is used to get the semantic vector representation for a given input utterance. We use Long Short Term Memory Recurrent Neural Network (LSTM-RNN) as an encoder. LSTM reads the inputs in the forward direction and accumulates rich semantic information. As a complete sentence passes through LSTM, its hidden layer stores the representation for the entire input sentence (Palangi et al., 2015). This sentence representation is used classification.

We use a fully connected layer followed by a softmax layer for classification. The fully connected layer learns function from sentence representation fed to it by the LSTM layer. Softmax layer gives the output probabilities for intent labels. We have illustrated this architecture using figure 3.

## 4 Experimental Results

In this section, we share the details of the benchmark and custom datasets, describe the training set-up, present experimental results, and compare our model with existing baselines.

### 4.1 Datasets

To compare our model with existing approaches, we benchmark it against two public data sets. First is the widely used ATIS dataset (Hemphill et al., 1990), which contains audio recordings of flight reservations. The second dataset is the custom-intent-engines dataset called SNIPS (Coucke et al., 2018) which is collected by Snips voice assistant. Details about both the datasets can be found in table 1. SNIPS dataset is more complex as compared to ATIS dataset because of multi-domain intents and relatively large vocabulary. We use the datasets that are pre-processed by Goo et al. (2018) with the same partition for train, test, and validation set.

**Custom dataset:** We have also curated a custom dataset. For initial data creation, we use user trial and scrape webpages. We define 10 intent labels to annotate custom dataset. These intent labels and their examples are given in table 2. However, the size of the data that is collected is not sufficient to train a neural network. Therefore we use different data augmentation techniques to increase data size. We use transformer-based data augmentation followed by synonyms replacement. In transformer-based data augmentation, we fine-tune a pre-trained BERT (Devlin et al., 2018) model on collected data. We use this fine-tuned BERT for making predictions on a huge unlabeled corpus. We only consider sentences that are classified with high confidence. We verify these sentences to ensure that predictions by BERT are correct. Further, we use synonyms replacement for augmentation. In a sentence, we randomly replace 30% of the words with their synonyms. The final data distribution is given in figure 4. For testing, we have curated a set of 100 sentences manually.

| Attributes | ATIS | SNIPS |
|---|---|---|
| No. of intents | 21 | 7 |
| Vocabulary Size | 722 | 11241 |
| Train set size | 4478 | 13084 |
| Test set size | 893 | 700 |
| Validation set size | 500 | 700 |

Table 1: Details of benchmark datasets.

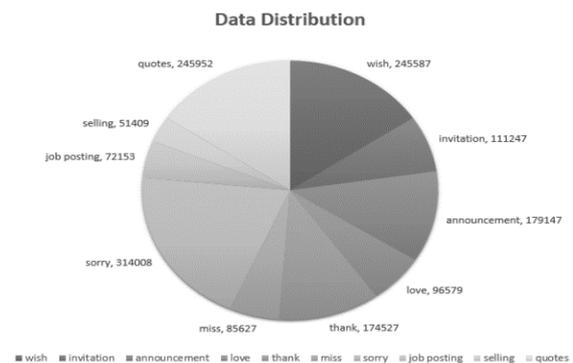

Figure 4: Data distribution in custom dataset

### 4.2 Training

We use the same set of parameters for training model on both benchmark datasets. We fix maximum sequence length to 25. We initialize word embedding with 50 dimensional pre-trained GloVe embedding. Character embedding size is set to 15. Kernel sizes are set to 3, 4, and 5 and filter sizes are set to 10, 20, and 30 in 3 convolution layers. LSTM layer has 128 units. Categorical cross-entropy is used for loss computation & Adam optimizer (Kingma and Ba, 2014) is used to minimize loss. The batch size is set to 16. Constant learning rate of 0.001 is used. Models are trained for 10 epochs.

Following Goo et al. (2018) we use accuracy as the metric for the evaluation. After each epoch, we evaluate the performance of the model on validation set. The model performing best on validation set is then evaluated on test set. To address the issue of random initialization, we repeat this process 20 times and consider the average accuracy for analysis.

For custom dataset, we use only 50K sentences per intent label. We also limit word vocabulary size to 12K most frequent words. Batch size is set to 64. Rest all parameters and hyper-parameters remains same as training benchmark datasets.

### 4.3 On-device deployment

We use TensorFlow (Abadi et al., 2016) to build all our models. We use Tensorflow Lite (tflite) to support on-device execution. The trained models on SNIPS and ATIS datasets are converted to tflite format. The size of the models is reduced by post-training quantization. The final size of our models trained on ATIS and SNIPS is 172 KB and 686 KB respectively. The size of the model trained on custom data set is 786 KB.

| Model | ATIS | SNIPS |
|---|---|---|
| Joint Seq (Hakkani-Tur et al., 2016) | 92.6 | 96.9 |
| Attention BiRNN (Liu and Lane, 2016) | 91.1 | 96.7 |
| Slot-Gated Full Atten (Goo et al., 2018) | 93.6 | 97.0 |
| Slot-Gated Intent Atten (Goo et al., 2018) | 94.1 | 96.8 |
| Self-Attentive Model (Li et al., 2018) | 96.8 | 97.5 |
| Bi-Model (Wang et al., 2018) | 96.4 | 97.2 |
| CAPSULE-NLU (Zhang et al., 2019) | 95.0 | 97.3 |
| SF-ID Network (E et al., 2019) | 96.6 | 97.0 |
| Stack-Propagation (Libo et al., 2019) | 96.9 | 98.0 |
| Stack-Propagation + BERT (Qin et al., 2019) | 97.5 | 99 |
| Our Model | 99.53 | 98.95 |

Table 3: Performance of our model compared to current existing approaches

### 4.4 Baselines

We compare our model with existing deep learning based models for intent classification. They are as follows: Joint Seq (Hakkani-Tur et al., 2016), Attention BiRNN (Liu and Lane, 2016), Slot-Gated Attention (Goo et al., 2018), Self-Attentive Model (Li et al., 2018), Bi-Model (Wang et al., 2018), Capsule-NLU (Zhang et al., 2019), SF-ID network (E et al., 2019) and Stack-Propagation (Qin et al., 2019).

| Metrics | ATIS | SNIPS |
|---|---|---|
| Model Size | 172 KB | 686 KB |
| Inference Time | 1.87 ms | 1.9 ms |
| RAM | 4850 KB | 4822 KB |

Table 4: On-device model performance

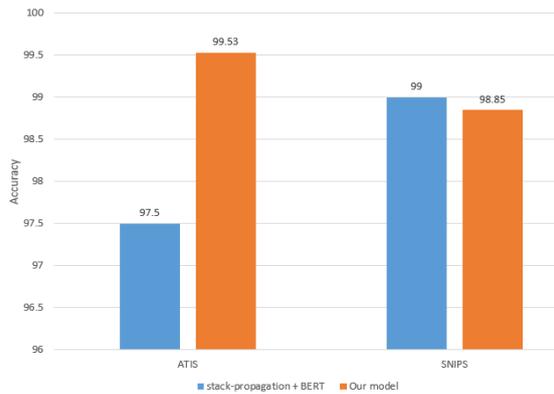

Figure 5: Accuracy comparison of our model with current state of the art

For all the baselines, we utilize the results reported by Qin et al. (2019).

### 4.5 Results

Detailed performance comparison of our model with existing approaches is shown in table 3. Our model achieves an average accuracy of 98.95% and 99.53% on SNIPS and ATIS datasets respectively. The variance of 0.026 on SNIPS and 0.01 on ATIS datasets prove the robustness of our model. Our model outperforms the current state-of-the-art Stack Propagation framework + BERT by 2.03% on ATIS dataset. On SNIPS dataset, our model achieves results comparable to the state-of-the-art and outperforms all other approaches. It is worth noticing that our model has much fewer parameters as compared to the state-of-the-art model. The model trained on custom dataset achieves 98% accuracy on custom test set.

We also measure the system-centric metrics for our models which are presented in table 4. We use a Samsung galaxy A51 device (4 GB RAM, 128 GB ROM, Android 11, Exynos 9611) for these experiments. Inference time includes the time required to pre-process the input text, tokenization, model execution, and label determination. We infer complete test set of datasets on device and report its average inference time. As stated in table 4, our models have an inference time of ~2 milliseconds. We also report maximum RAM usage during on-device inferencing is less than 5 MB.

All the above-mentioned results prove that our model is not only accurate but it also has low inference time and RAM usage. This proves that our model is efficient for running on low-resource devices.

## 5 Conclusion

In this paper, we present an on-device intent classification architecture that uses character level features to enrich the word representation. Our experiments prove the effectiveness of our model as it achieves state-of-the-art results on benchmark datasets. System centric metrics like RAM usage and inference time shows that our model is fast and light-weight to be deployed on low-resource devices. For future work, we want to extend this approach for slot filling and experiment with a joint model for on-device intent detection and slot filling.